# Teaching Robots to Build Simulations of Themselves

**Authors:** Yuhang Hu[1]*, Jiong Lin[1], and Hod Lipson[1,2]*

**Affiliations:**

[1]Creative Machines Laboratory, Mechanical Engineering Department, Columbia University, New York, NY 10027, USA

[2]Data Science Institute, Columbia University, New York, NY, 10027, USA

*Corresponding author. Email: yuhang.hu@columbia.edu, hod.lipson@columbia.edu

**Abstract:**

The emergence of vision catalyzed a pivotal evolutionary advancement, enabling organisms not only to perceive but also to interact intelligently with their environment. This transformation is mirrored by the evolution of robotic systems, where the ability to leverage vision to simulate and predict their own dynamics marks a leap towards autonomy and self-awareness. Humans utilize vision to record experiences and internally simulate potential actions. For example, we can imagine that, if we stand up and raise our arms, the body will form a 'T' shape without physical movement. Similarly, simulation allows robots to plan and predict the outcomes of potential actions without execution. Here we introduce a self-supervised learning framework to enable robots to model and predict their morphology, kinematics and motor control using only brief raw video data, eliminating the need for extensive real-world data collection and kinematic priors. By observing their own movements, akin to humans watching their reflection in a mirror, robots learn an ability to simulate themselves and predict their spatial motion for various tasks. Our results demonstrate that this self-learned simulation not only enables accurate motion planning but also allows the robot to detect abnormalities and recover from damage.

**One-Sentence Summary:**

A learning algorithm allows robots to model their own 3D morphologies, kinematics, and motor control from a 2D video for task planning.

**Main Text:**

# INTRODUCTION

Every robot begins its life in simulation. Long before any robot is ever physically built, engineers meticulously design computer-aided design (CAD) models, craft kinematic equations, and estimate a variety of physical coefficients describing the robot to be. Using this simulation, the robot can be tested, programmed, and taught a variety of tasks. Only once this simulated robot meets intended performance, is it physically built and operated. Simulations remain essential for robot learning because learning in real life is slow, energetically expensive, and risky [1, 2, 3, 4].

Humans and animals also use a simulator to learn. In contrast to robots, however, human and animal self-simulation is learned and evolved, not programmed by a designer. A child, for instance, creates a mental model of its own body by observing itself directly (or through a mirror reflection), refining its movements and gaining a deeper understanding of its body kinematics and motor control in relation to the environment. While we aim to imbue robots with a similar level of morphological self-awareness and adaptability, it's important to recognize the current limitations in robotic technology compared to the complex cognitive processes observed in living beings [5].

Body self-awareness, as demonstrated through mirror self-recognition, remains a unique cognitive ability reserved for only a few species, such as humans, chimpanzees, and orangutans. Most primates, despite prolonged exposure to mirrors, continue to perceive their reflection as another individual rather than recognizing it as an image of themselves [6, 7, 8]. This complex cognitive capacity to process mirrored information about oneself, which most primates seem to lack, raises pertinent questions about the nature of consciousness, the emergence of self-awareness, and its link to the development of the mind. In the human context, mirrors serve as instrumental tools for cognitive development, enhancing skills, refining gestures, and enabling a deeper understanding of one's body in relation to the surrounding environment. [9, 10].

In robotics, understanding their own 3D morphology is becoming increasingly essential for tasks such as motion planning and navigation [11, 12, 13, 14, 15]. This not only enables them to execute comprehensive planning tasks but also enhances their adaptability. Previous work in robotic self-modeling only focused on predicting the center of mass or the end-effector position of robots [16, 17]. Notable contributions include Vaughan and Zuluaga's demonstration of sensorimotor self-simulation for navigation planning with incomplete knowledge [18], Wittmeier et al.'s development of a compliant humanoid robot torso using reinforcement learning for object-reaching tasks [19] and Blum et al.'s use of simulation-based internal models for safe navigation in crowded environments [20]. Recent advancements in computational self-modeling have primarily leveraged several depth cameras to capture and predict a robot's kinematics [21]. The Neural 3D Visual Self-Model's groundbreaking potential highlights its pivotal role in 3D motion planning and control tasks. More than just a morphology prediction, this visual self-model exhibited handling damage detection, identification, and subsequent recovery for real-world robotic applications.

Previous work, however, relied on five well-calibrated depth cameras, necessitating complex setups and intricate calibration processes. Furthermore, fusing multiple depth images from different devices into a coherent point cloud introduces noise and potential inaccuracies. The question we aim to address in this aper is as follows: Can we teach robots to learn its body morphology, kinematics, and motor control based on a single 2D camera, just as humans look into the mirror to learn without leveraging synthesized 3D ground truth data for supervision? Enter Neural Radiance Fields (NeRF), a groundbreaking approach in Computer Vision [22, 23, 24, 25, 26, 27, 28, 29, 30]. NeRF models exploit volume rendering combined with explicit neural scene representation, trained through multi-view images devoid of 3D or depth supervisory information.

In this paper, we leverage such ability from NeRF to advance the boundaries of this research with a focus on modeling the robot's morphologies, kinematics, and motor control within a continuous 3D domain. We propose a self-supervised learning framework enabling robots to leverage videos for self-modeling, akin to a utilize mirrors to understand and refine their movements (Fig. 1A). We introduce Free Form Kinematic Self-Model (FFKSM) which is a query-based model that can output the robot occupancy in 3D space (Fig. 1C). By keeping both the spatial coordinates and joint angles continuous, we can probe the robot's 3D morphology and kinematics with precision at any spatial location using any joint angles in action space. Besides, the model is fully differentiable, and we can search the robot motor commands using gradient-based optimization in real time. We also show how to use this model to perform basic manipulation tasks: 3D morphology prediction, motion planning without relying on kinematic equations, and abnormal detection alongside damage recovery (Fig. 1B). Notably, the model is highly efficient, requiring only 333 KB of memory—significantly less than the 1.1 MB required by previous approaches. This substantial reduction in memory usage underscores the model's enhanced efficiency.

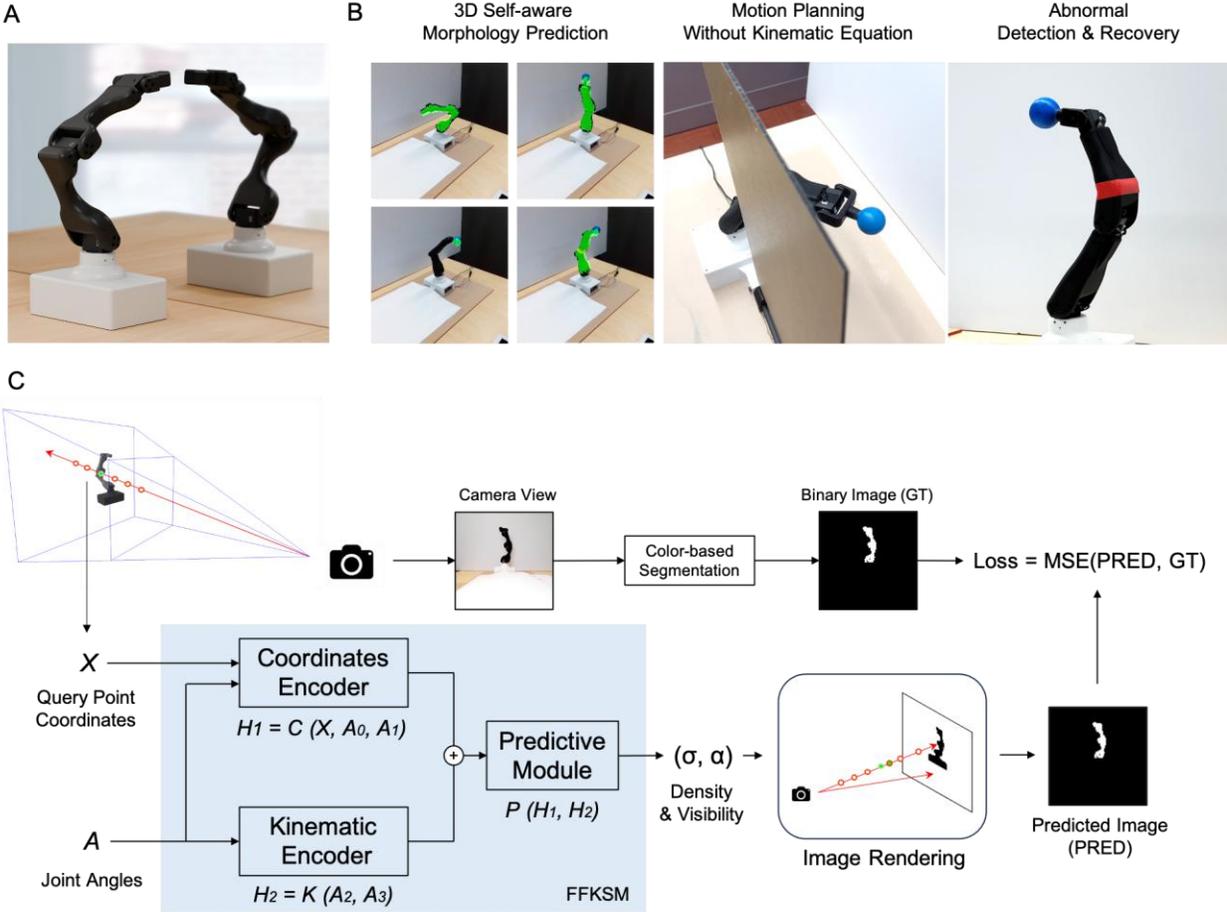

**Fig. 1**: **Entire pipeline overview:** **(A)** An illustrative representation of our foundational concept: a robot looking into a mirror. trying to build a model of itself by moving its body to observe changes. **(B)** Leveraging the model, the robot can predict its own morphology and perform a variety of manipulation tasks. **(C)** Free Form Kinematic Self-Model representation. The model comprises three deep neural networks: a coordinates encoder, a kinematic encoder, and a predictive module. By processing 3D point coordinates and joint angles, it predicts the density and visibility of queried points. This information is then used to render a predicted image, which is compared with a segmented binary image for training.

## RESULTS

### 3D Morphology Predictions Using FFKSM

We conduct experiments both in the simulation and the real-world, using two 4 degrees of freedom robot arm. During data collection, as illustrated in Figure 2A, we discretized the action space using a 9-degree interval and sampling after traversing the entire action space, allowing for a mix of small and large perturbations. The camera saved the video frame once the motor encoder detects the joint angles met the motor commands. Each video frame is an RGB image, with a size of $100 \times 100$. In the data collection phase (Supplementary Video 1), we collected a total of 12,000 data points. Among them, 10,000 are allocated for model training and validation at an 8:2 ratio, while the remaining 2,000 are for testing purposes.

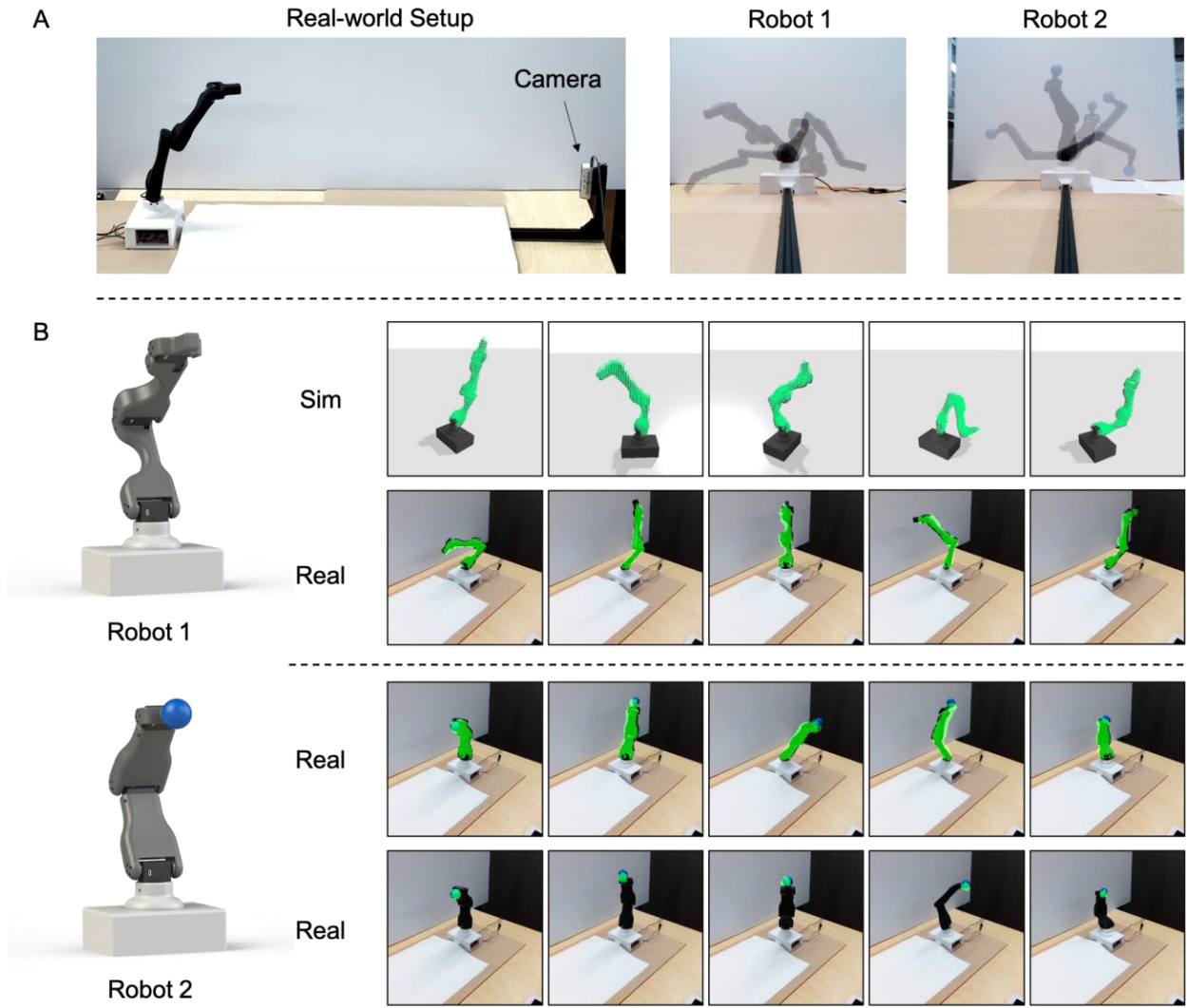

**Fig. 2: Real-world experimental setup and 3D morphology prediction.** **(A)** Real-world experimental setup. While the robot is doing motor babbling, the camera capturing its motion (left). The synthesized images represent typical captures by the camera (right). **(B)** Robot 1 and Robot 2 Hardware (left). On the right, predictions of robot 3D morphology using the free-form kinematic self-model are shown. The images demonstrate simulation and real-world predictions for both robots and the end-effector. Within each image, the actual robot is depicted in black, while predicted non-zero density areas are highlighted in green. Please see Supplementary Video 2 and Supplementary Video 4 for more information.

In Fig.2 (B) we show the robot 3D morphology predictions obtained through the free-form kinematic self-model. The predicted results are shown as green points in both simulation and real world, while the robot ground truth morphology is shown in black. The green points in the 3D space stand for the queried points where the model estimate it is occupied by the robot. Upon comparative analysis of the outcomes from the first and second rows, the results demonstrate a consistent performance, underscoring the robustness of our method in real-world scenarios. In the third row, we show that our model can effectively predict robot 2 morphologies which have different kinematics. This highlights the adaptability of our approach, making it suitable for diverse robots without the need for task-specific adjustments.

Beyond predicting diverse robot morphologies, our model can also predict different components of a given robot. For example, to predict the end-effector of robot 2, we can achieve it through two training processes. During the first training phase, we train the model to predict the entire morphology of robot 2. Subsequently, we use video frames only that have the robot end-effector to fine-tune the model. Such images are procured by adjusting the color filter of the color-based segmentation module to isolate the blue pixels. The results demonstrate that our method can train models which have a capability to accurately predict robot occupancy in 3D space. This skill extends across different robots and even to specific robot components.

**Motion Planning Using FFKSM**
In this section, we illustrate how to use the trained model to achieve motion planning tasks. The FFKSM can predict robot morphology in 3D space and also understand the specific component occupancy as we present in the previous section. Here we use robot 2 with a ball head end-effector to perform two manipulation tasks (Fig. 3).

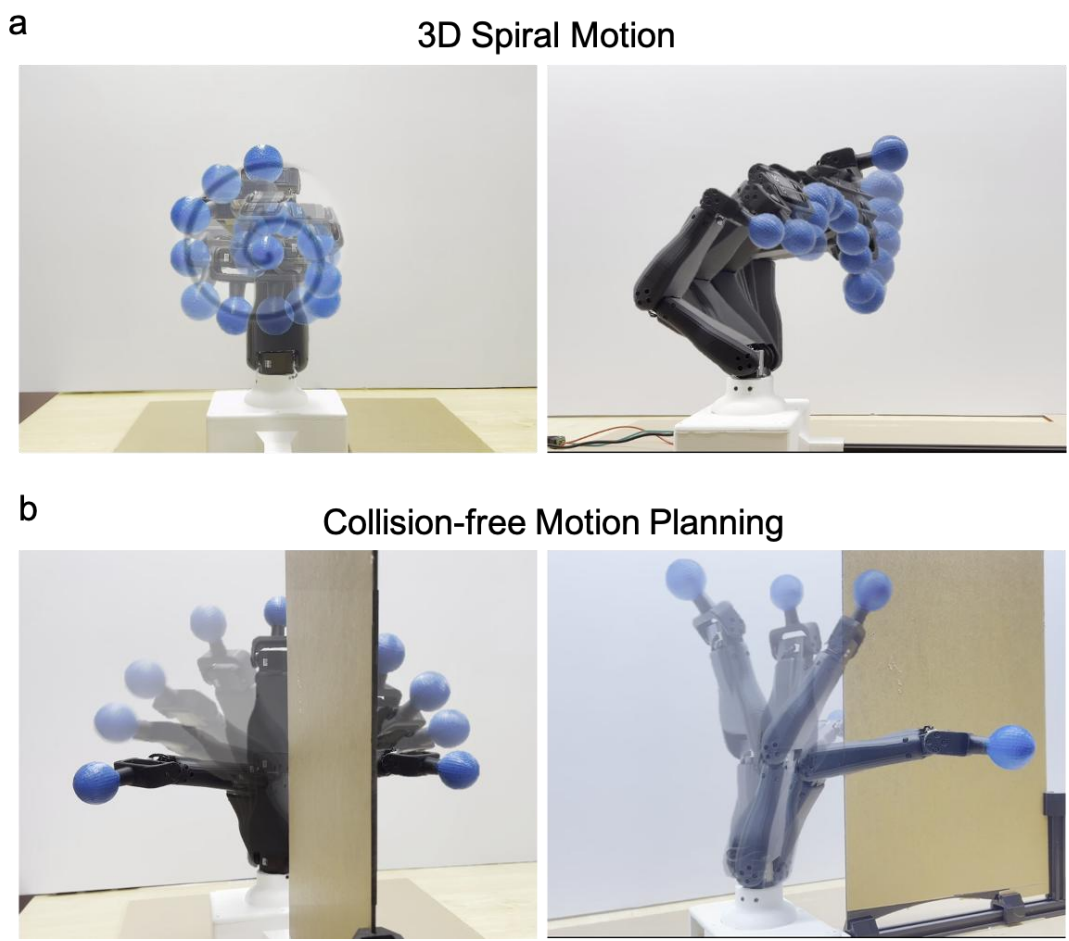

**Fig. 3: Real Robot 2 performs motion planning tasks using FFKSM. (a)** The robot tracks a 3D spiral trajectory. The images generated from a video illustrate the robot spiral trajectory. **(b)** The robot performs a collision-free motion planning task. The pictures show the robot's motion from a starting point to its target, navigating around obstacles to ensure no collisions.

The first task involves controlling the robot end-effector along a 3D spiral trajectory. Unlike simple linear or planar movements, a spiral trajectory inherently involves three-dimensional space. This allows for the demonstration of the robot's ability to operate in a 3D environment, which is crucial for many real-world applications. This task is equivalent to performing Inverse Kinematic (IK) control, where the objective is to determine the joint angles based on the end-effector position. Traditional approaches would necessitate deriving equations rooted in the robot's mechanical geometries. In contrast, our method circumvents this necessity, permitting us to utilize gradient-based optimization to realize IK control. This is achieved without prior knowledge of the robot's hardware specifics or its Forward Kinematics (FK). Given that the 3D query points, joint angles, and our model are all differentiable, we present an algorithm that optimizes the joint angles by minimizing the distance between the predicted end-effector location and the target joint position.

To solve this problem, we use a model that is fine-tuned with end-effector data which can predict the 3D points occupied by end-effector. The end-effector position can be defined as the average position of the query points with non-zero density. Assume Q is the set of all query points, and $Q_{occupied}$ is the subset of those points with non-zero density, i.e.,

$$Q_{occupied} = \{q \in Q | \sigma(q) > 0\}$$

Where $\sigma$ is the density of query point as predicted by the model. Let $\theta$ represents the joint angles of the robot s.t.: $-90° \leq \theta_i \leq 90°$ for i = 1, 2, 3, 4. The end-effector position, $p_{ee}$, is the average position of all the points in $Q_{occupied}$:

$$p_{ee}(\theta) = \frac{1}{|Q_{occupied}|} \sum_{q \in Q_{occupied}} q$$

During the optimization process, we utilize the Adam optimizer for efficient convergence, with a learning rate set at 0.04 [31]. Backpropagation ensures iterative refinement of the joint angles by minimizing the Mean Squared Error (MSE) loss.

The second task is navigating the robot end-effector from a starting point to a target point without any collisions. We assume that the positions and dimensions of any obstacles in the robot's path are known a priori. Our approach leverages two FFKSMs. The first predicts the entirety of the robot's morphology and is used for collision detection. The second predicts the position of the robot's end effector, serving as a heuristic calculator.

Through the model that can predict the whole-body morphology, the robot can estimate its body occupancy in a given configuration. A collision is deemed to occur if any portion of the robot's predicted morphology, represented by query points with non-zero density, intersects with the predefined obstacle region. To generate the robot collision-free trajectory, we use the Rapidly exploring random tree (RRT) algorithm [32]. RRT, known for its efficacy in high-dimensional spaces. The model predicting the end effector position, provides heuristics and guides the trajectory generation process. By leveraging two FFKSM, the robot can achieve collision-free motion planning. Please see Supplementary Video 4 for more information.

## Abnormality Detection and Damage Recovery

In the evolving landscape of autonomous robotics, the ability of robots to adapt to physical changes or damage in real-world settings is paramount. This adaptability is particularly crucial in scenarios where human intervention is impractical or impossible, such as in hazardous environments, healthcare, manufacturing, or domestic applications. Traditional models, relying on static, unchanging morphologies, fall short in these dynamic conditions. Our study addresses this gap by focusing on a robot's ability to recognize and adapt to morphological changes, such as damage or wear and tear, enhancing its real-world applicability and longevity.

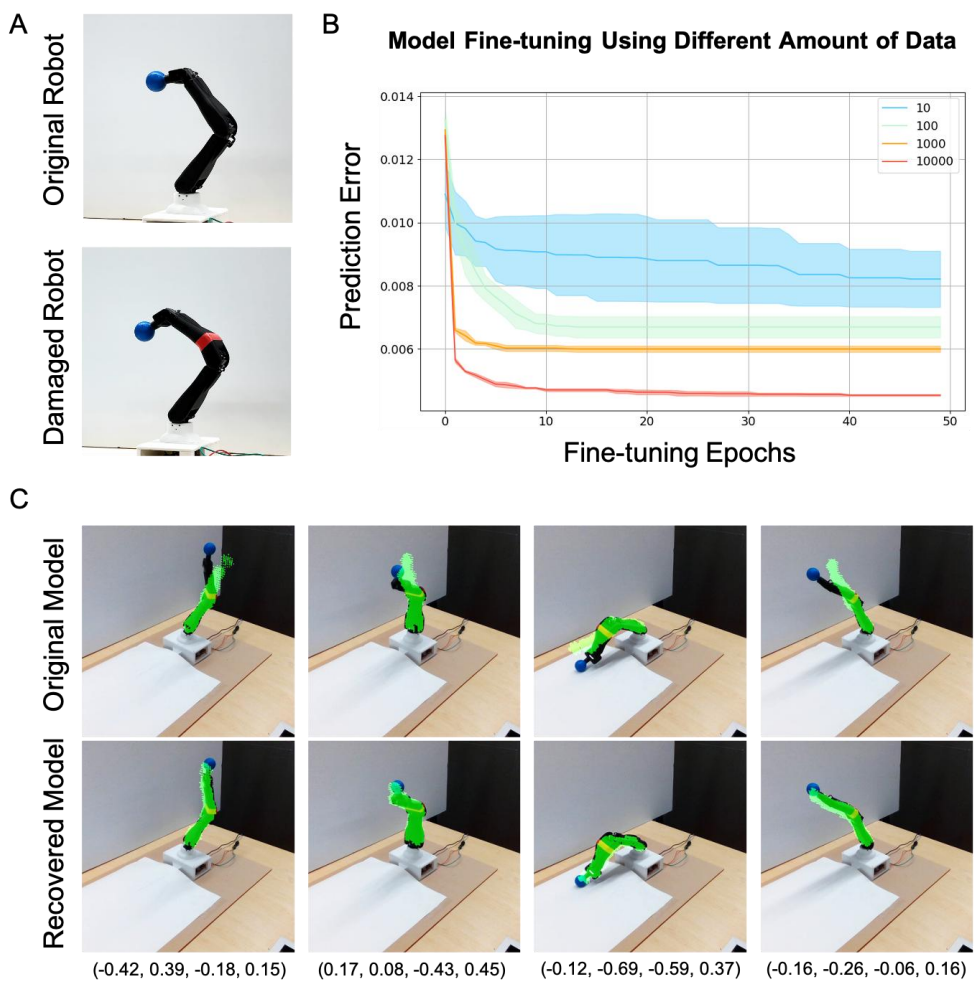

**Fig. 4: Recovery of Robot Kinematics After Damage.** **(A)** Pictures of the actual robot arm 2 used in the experiments. The bending area marked in red to simulate damage. **(B)** A plot indicating the divergence between the model's predictions and the ground truth during fine-tuning. The plot encompasses data from four distinct quantities, showcasing the model's adaptability with varying levels of post-damage data. **(C)** Comparative visualization of the robot's morphology predictions before and after retraining with post-damage data. The green overlay on the real robot arm indicates the predictions of our model. This comparison highlights the robot recovers predictive accuracy after retraining.

The versatility of the Free-Form Kinematic Self-Model extends beyond mere morphology prediction and motion planning. It can detect abnormalities and facilitate damage recovery. Our experiments utilized robot arm 2 with an RGB camera. Considering a scenario where the robot faces an overload, leading to a bent in its link. To simulate real-world damage, we use a bending robot component which mirrors common types of damage robots might encounter in various operational environments, providing a realistic basis for our experiments. We use 3D printing to fabricate the damaged component and integrate it with Robot 2 as shown in Fig. 4(A). This mechanical deformation induces a noticeable disparity between the images captured by the real camera and the 2D image predictions from the model, as seen in the first row in Fig. 4(C).

The recovery (fine-tuning) phase involved retraining the model using the post-damage data. This process was critical in enabling the robot to rebuild its FFKSM to align with its new physical state. With the additional real-world data after the robot damage, the robot refines its predictions to align with the deformed 3D morphology. This adaptation is quantified in Fig.4 (C), which plots the decreasing divergence between the model's predictions and the ground truth as it is fine-tuned with the post-damage dataset. The plot shows the results from varied data quantities (10, 100, 1000 and 10000), demonstrating the model recovery effect at different data scales.

The results shown in Fig.4 (C) demonstrate the performance in detecting and recovering from damage. This recovery ability advances the field of robotic self-modeling, particularly in the context of autonomous adaptation to physical damages. This has profound implications for the future of robotics, especially in applications requiring high degrees of autonomy and resilience.

**Quantitative Evaluation of 2D Image Prediction**
To assess the performance of our model, we use 2D predicted images, since directly measuring real-world 3D morphology without other sensors or equipment is unavailable. We use a test dataset comprising 2000 samples, unseen during the model training phase. Through querying 64 random points along each ray that spans robot space, our image rendering module can generate predicted images. The Mean Squared Error (MSE) serves as our metric to gauge the difference between these predictions and the ground truth.

Our experiments span three distinct configurations in a real-world environment: Full morphology prediction of Robot 1, Full morphology prediction of Robot 2, and Prediction of only the end-effector for Robot 2. We compare our method (OM) to the other two baseline methods. The first baseline, Random Selection (RS), arbitrarily picks an image from the training dataset. These randomly chosen images share a distribution similar to our test dataset, so we use this as one of our baselines. The second baseline, Nearest Neighbor (NN), chooses the image that has minimum L2 distance, in terms of joint angles, to the ground truth. The purpose of designing this baseline is that robotic arms with similar joint angles will have similar images from the camera view. Figure 5 shows that our method can accurately predict robot morphology and has lowest errors compared to the baselines across all tested configurations. Figure 6 presents the hardware details of the robots used in our experiments.

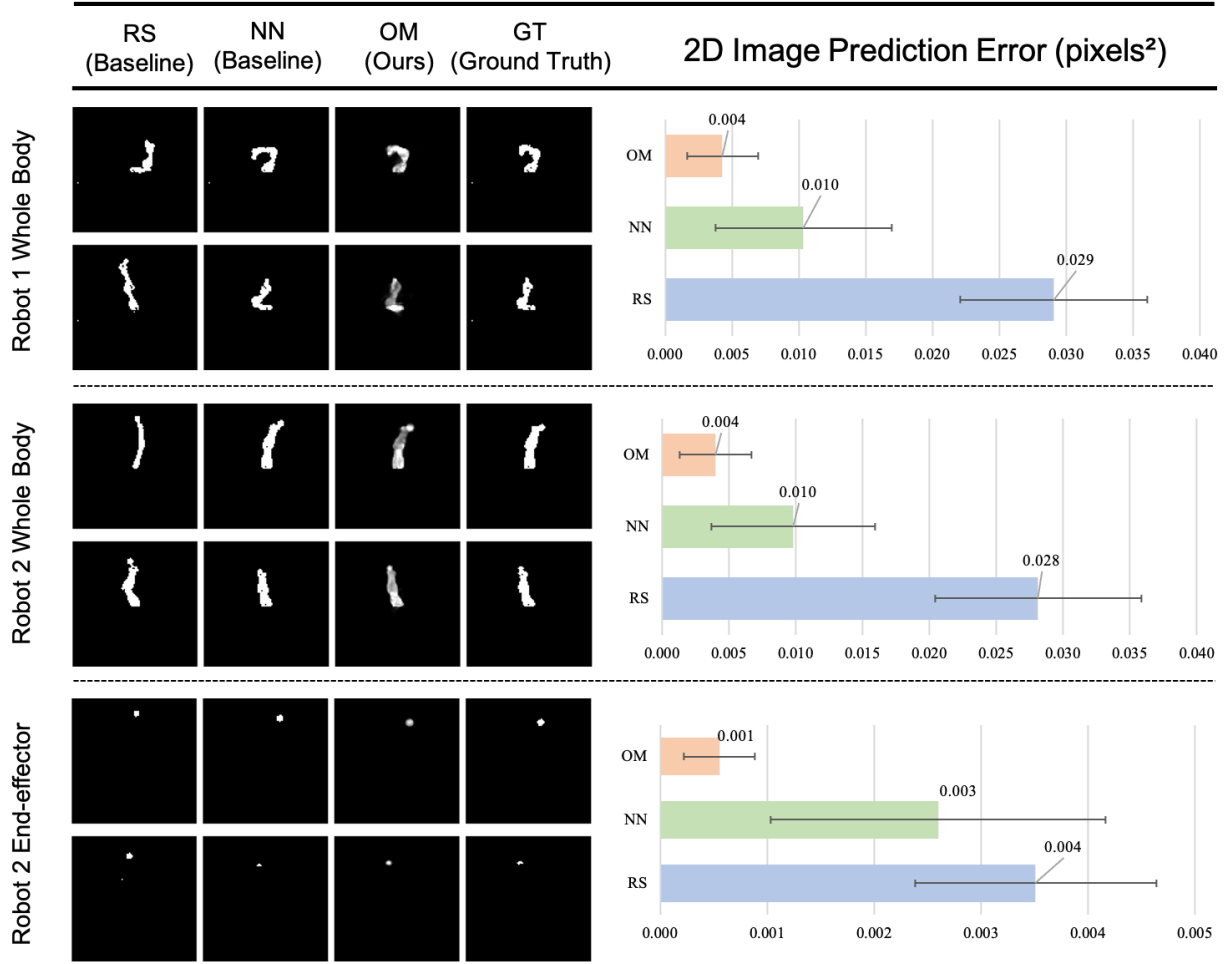

**Fig. 5: Comparison of Predicted Robot Morphology and Quantitative Results.** The left side presents a side-by-side comparison of the robot morphology predicted by baseline methods, our proposed method, and the ground truth, demonstrating the accuracy of our method in capturing the robot's shape. This prediction error is quantified using mean squared error (MSE) between the predicted robot images and the ground truth (GT), measured in units of squared pixels. The right panel showcases three bar plots illustrating quantitative results for a sample size of n=2000: our method consistently demonstrates lower prediction errors and standard deviations across all tests. The error bars in the bar plots are presented as mean values ± standard deviation (SD).

## DISCUSSION

In summary, our experiments of 3D motion planning demonstrate the capability of the Free Form Kinematic Self-Model in addressing robotic manipulation tasks without the traditional reliance on kinematic functions or CAD models. Recent methods rely on intricate setups, necessitating the use of multiple depth cameras and laborious calibration processes to create ground truth for the model. However, our approach emphasizes the capability of a robot to self-model its morphologies, kinematics, and motor control through 2D visual feedback. This visual feedback loop is analogous to a human using a mirror, providing fast adaptation and resilience. Our approach not only facilitates morphology predictions and motion planning but also offers adaptability to a diverse range of robotic platforms. Our method bypasses the challenges encountered in manual modeling, particularly in cases where robots possess complex, evolving or damaged structures.

The versatility of the FFKSM extends to its potential application in compliant and mobile robots. For mobile robots, our model facilitates on-the-fly kinematic recovery, essential for operations in hazardous environments such as nuclear disaster sites. When such a robot encounters physical deformations, the FFKSM can automatically recover the robot's functionalities using visual feedback from onboard cameras (monitoring robot arm), thus ensuring continued operation despite damages.

Regarding compliant robots, which often involve complex interactions with their environment due to their flexible structures, the FFKSM offers a promising direction for future research. These robots pose unique challenges in kinematic modeling due to their variable morphology under different loads. Our future work will explore integrating sensors that measure joint torque or motor current to enhance the FFKSM's applicability to compliant robotics. This will involve developing adaptive models that can dynamically adjust to the soft robot's changing state, providing a more accurate modeling of its morphologies and movements.

Through the experiments we demonstrated that our model has proficiently internalized the intricacies of robot kinematics, motor control and its 3D morphology. This foundational understanding can be harnessed effectively to navigate and address real-world challenges, setting a new paradigm in robotic motion planning and control.

Our key contributions can be summarized below:

1. Robustness and Generalizability: Our method displayed consistent performance in both simulated and real-world environments. This proves the robustness of the approach, making it reliable in varied contexts. Additionally, our approach could be extended across different robots through its generalizability.
2. Versatility of Application: Beyond mere kinematic modeling and morphology synthesizing, our method showcased its adaptability in abnormality detection and recovery processes. In real-world scenarios where robots may experience wear, tear, or damage, such a feature can recover the robot without human intervention.
3. Tackling Motion Planning Tasks: Traditional motion planning tasks, which rely on explicit kinematic equations and CAD-based modeling, can be computational-consuming and inflexible in dynamic environments. The Free Form Kinematic Self-Model method, through its gradient-based optimization and adaptability without human-predefined kinematic equations, provides an alternative approach that's more adaptable and autonomous.
4. Real-world Applicability: A quintessential feature of our research was its emphasis on real-world applications. Whether it was in the prediction of morphologies, motion planning, or damage recovery, our approach consistently aligned with real-world scenario.

A potential limitation of our work, however, is the reliance on 2D images for quantitative evaluations. Given that direct measurements of real-world 3D morphology without auxiliary equipment is currently unattainable, our quantitative evaluation of 3D prediction is a compromise. In future research, enhancing the camera resolution and employing dynamic sampling techniques can substantially improve the model's accuracy. Additionally, integrating 3D modeling sensors could provide a comprehensive evaluation of the robot's morphology.

In conclusion, the strides taken in this paper mark a notable progression in the domain of robotic self-modeling. The potential applications are vast, from autonomous robotics in industrial settings to those deployed in challenging terrains and environments. We believe that the principles laid out here will not only facilitate the rapid development and deployment of robotic solutions but also pave the way for robots that are more adaptive, resilient, and efficient.

## Methods

### Free-Form Kinematic Self-model

As autonomous robotics advance, the need for robots capable of minimal human supervision becomes paramount. Robots should learn to create models for themselves that not only generate motor commands to solve the problems but can adapt to changes ang recovery from damage [33]. Engineers simplified the robot morphology and based on rigid body assumption to design a specific kinematic function for a robot [34]. To achieve morphology simulation for collision-free planning, it requires a detailed CAD model. In most simulations, the simple non-convex morphology, such as collision boxes is used to represent the robot morphology. As robots become more complex, manually modifying these functions when robot structures change becomes less feasible and less adaptable. For example, in instances where a robot sustains damage, the ability to quickly self-recover is also crucial.

To address these challenges, we introduce a visual self-supervised learning approach that allows the robot to autonomously learn a FFKSM including robotic morphology, kinematics, and motor control from video. The FFKSM is query-based deep neural networks trained by videos through motor babbling, so the model is task-agnostic and requires minimal human supervision. As depicted in Fig.1C, a robot is positioned within the camera view. By inputting specific 3D point location $X \in \mathbb{R}^3$, including the robot arm joint angles $A \in \mathbb{R}^4$, the model can predict whether the 3D point location is occupied by the robot and whether it is visible through the camera.

The FFKSM is composed of three deep neural networks: a coordinates encoder, a kinematic encoder, and a predictive module. The Coordinates Encoder takes in world coordinates X and a projection matrix $T$ computed through the first two joint angles $A_0$, $A_1$. This transformation produces virtual coordinates $X'$ in which gives an equivalent effectiveness that the camera is the entity in motion and the first two robot arm joints are static. Given $R_{yaw}(A_0)$ and $R_{pitch}(A_1)$ are the rotation matrices for rotations around the robot first two joint axes respectively, the transformation matrix $T$ is the product of these rotations, represented mathematically as:

$$X' = T^{-1}X = (R_{pitch}(A_1)R_{yaw}(A_0))^{-1}X$$

Then the virtual $X'$ is fed into the Coordinates Encoder. Subsequently, the Kinematic Encoder processes the remaining joint angles. Concatenating the output of the Coordinates Encoder $C(\cdot)$ and the output from the Kinematic Encoder $K(\cdot)$ as input, the Predictive Module $P(\cdot)$ estimates the density $\sigma_{ijk}$ and visibility $\alpha_{ijk}$ at position (i, j) for the ray k in the discretized 3D space. The model can be expressed as:

$$(\sigma_{ijk}\alpha_{ijk}) = P(C(X'_{ijk}), K(A_2, A_3))$$

However, to enhance the model's performance and effectively differentiate between density and visibility, we compute the density using the following function:

$$\sigma_{ijk} = 1 - \exp(-ReLU(B_{ijk}))$$

Here, $B_{ijk}$ is the second raw output of the predictive module for point k along ray ij, and ReLU is the rectified linear unit activation function. By employing this function, we ensure that density values are positive and smoothly increasing, which aids in stabilizing the training process and improving convergence.

To render a predicted image with length and width dimensions W x H, we query M points in 3D space along the ray projected from the camera. The model generates the density and visibility for each point. The image rendering process aggregates the product of density and visibility values for every point along the projected ray. Considering all the M query points on each ray, the pixel value $Pred_{ij}$ at position (i, j) for the ray k, is:

$$Pred_{ij} = \sum_{k=1}^{M} \sigma_{ijk} \alpha_{ijk}$$

We use an RGB camera to capture the video frames of the robot during motor babbling. We use a color-based segmentation method to convert the RGB frames into binary images that exclusively highlight the robot arm. The Mean Squared Error (MSE) between the model's predicted images and these binary images is utilized as the loss function to optimize the model's training. Given Ground Truth as GT, the loss function can be represented as:

$$\mathcal{L} = \frac{1}{W \times H} \sum_{i=1}^{W} \sum_{j=1}^{H} (Pred_{ij} - GT_{ij})^2$$

**Robot Hardware**
In this study, we designed robot arms with four degrees of freedom, as illustrated in Fig. 6. Each arm is built on a common base that comprises yaw and pitch axis joints. The positioning of these joints allows for a camera, aimed at their intersection, to effectively monitor their interaction, as depicted in Fig. 6(D, E). This arrangement, where a fixed camera captures the movement of the joint axes, is functionally analogous to a moving camera observing a stationary robot base.

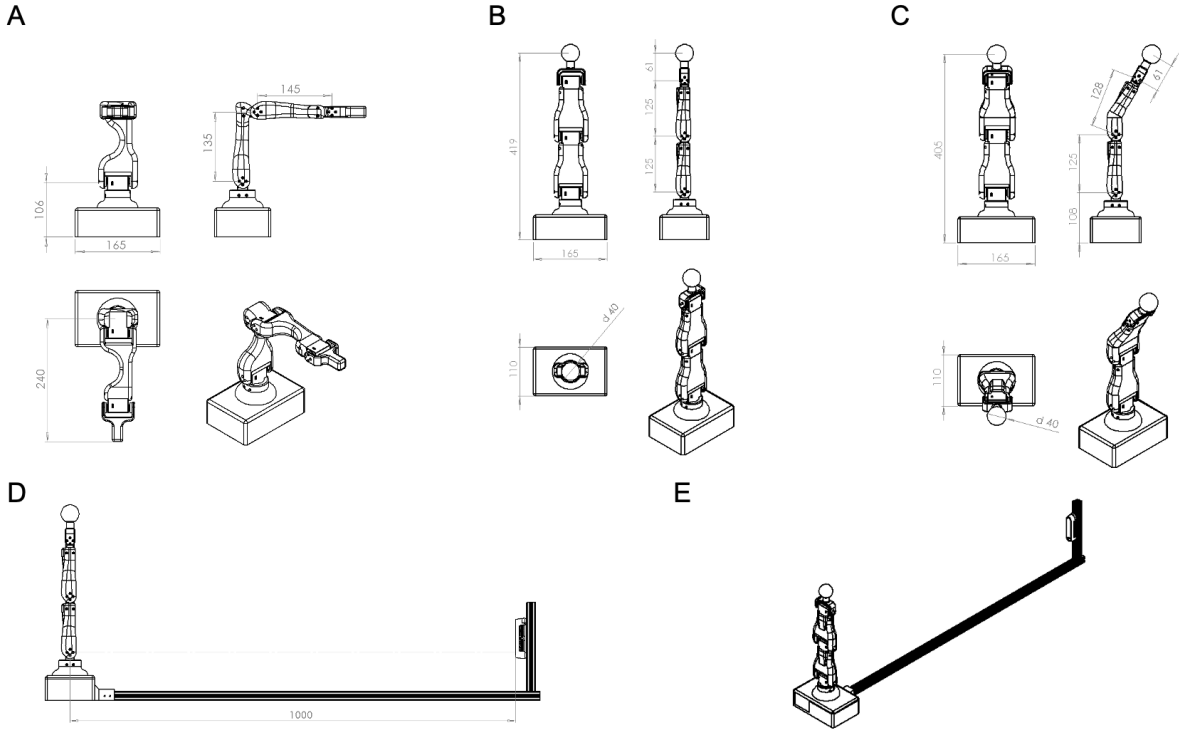

**Fig. 6: Mechanical Drawings of Robot Arms.** (A) Drawing of Robot Arm 1. (B) Drawing of Robot Arm 2. (C) Drawing of Robot Arm 3, featuring a bent second link compared to Robot Arm 2. (D, E) Robot Arm 2 equipped with a camera for data collection.

The primary distinctions between Robot Arm 1 and Robot Arm 2 lie in their morphological and kinematic differences. Notably, the third motor in Robot Arm 1 rotates 90 degrees relative to that in Robot Arm 2. Fig. 6(A, B, C) shows all joints in their neutral, zero-degree positions. Each joint is capable of a range of motion from -90 to +90 degrees. We select Serial Bus Servos with a torque of 25 KG·CM for the joints and use position control to actuate motors. An Nvidia Jetson Nano board serves as the onboard computer for processing. The fabrication of the robot arms was carried out using a Fused Deposition Modeling (FDM) 3D printer, utilizing Polylactic Acid (PLA) as the material.

For stable imaging, the camera is mounted to 2020 aluminum extrusions, ensuring consistent focus on the robot arm during its movements. The camera used is an Intel RealSense 435, configured for RGB imaging.

**Data Collection and Input Preparation**
In this section, we talk about the details of the data collection process. To efficiently collect data, we discretized the action space using a 9-degree interval and generated trajectories for the robot to execute all commands effectively. To ensure accuracy and avoid collisions, we implemented a feedback control mechanism. This involves using position encoders to monitor joint positions in real-time. When a joint reaches within 0.5 degrees of its target position, the camera captures and saves the frame. If a joint fails to reach this threshold within one second, we interpret this as a collision, halting movement in that direction. After collecting the dataset covering the entire action space, we randomly selected 10,000 data points for model training. This quantity was deemed sufficient based on preliminary tests. Each data point consists of a pair: the joint position and a corresponding image of the robot.

For capturing images, we positioned Robot Arm 2, which stands 439 mm tall, at a distance of 1 meter from the camera. This camera is equipped with a 42-degree field of view, ensuring comprehensive coverage of the robot. This setup ensures that the robot's full body is captured in the frame. For every image, joint angles remained consistent, and sampled a volume of $100 \times 100 \times 64$ points to query the model about their respective densities. This process took 0.3 second on a desktop with a single GPU (NIVIDIA GeForce RTX 3090). The first two rows present the results using Robot 1 hardware with simulation and real-world data. Pybullet [35] serves as our chosen physical simulation platform. Lacking direct ground truth for the 3D morphology, we employ the CAD model of the actual robot for comparative reference. To separate the robot from the background, we employed color-based segmentation. The robot arms were painted in black and blue, against a clean, uniform background. We further refined the background removal by calculating the median pixel values of each image. As the robot is the only moving element, the static background can be isolated using an image synthesized from these median values.

In our image processing, we normalized pixel values to the range [0,1]. A threshold of 0.15 in RGB channels was used to segment the robot body. For the blue end-effector on Robot Arms 2 and 3, a threshold of 0.4 was applied for pixel distance to the blue color. For this study, we resized images to 100×100 pixels, which yielded satisfactory predictive performance. Future enhancements with advanced equipment could allow for higher resolution images and more precise control. Extended Data Fig.1 visually represents the crucial steps in our data collection and image processing methodology, demonstrating the transition from raw visual data to processed binary images essential for the FFKSM training procedure.

In the training of the FFKSM, joint angles are obtained directly from the encoders. 3D position inputs are derived through a computational process that interprets the two-dimensional camera images into three-dimensional positions. This process utilizes the camera's point of view (POV), the fixed camera position relative to the robot, and the spatial location of image pixels. Essentially, we map the 2D data from the camera into a 3D space using a perspective transformation, which accounts for the camera's intrinsic parameters.

**Comparative Study with the original Neural Radiance Fields (NeRF)**
We have developed a model that generates two key outputs: **α** (density) and **σ** (visibility). This approach is crucial for accurately rendering images from a robotic perspective. When a camera captures the robot, only the side of the robot facing the camera affects the image pixels. Since we do not have direct ground truth data for density and rely on 2D images for loss computation, using density alone for image rendering is impractical. Instead, our model calculates the product of density and visibility to determine how an image is rendered. This means that if parts of the robot are not visible to the camera, their density does not influence the final image due to a visibility value of zero.

Our methodology shares parallels with NeRF, which employs a neural network to depict a scene and synthesizes new views from sparse 2D imagery. Like NeRF, our approach learns 3D information from 2D images, integrating spatial points as pixels to guide the model. But, distinct from NeRF, our data collection relies on a stationary camera, with the robot's primary joints manipulated to emulate varied camera perspectives. Whereas NeRF generates four outputs (RGB and volume density), our emphasis centers on space occupancy. In this section, we compare our method with two baselines using the original NeRF rendering methods. First, we explored the potential of adopting NeRF's volume rendering technique (VR), questioning if a singular output—volume density—might suffice for our model. In the VR rendering method, we calculate the density (α) as a function of the one of the raw outputs. One of the raw outputs for density can be represented as $\text{raw}_{density}$. The equation for calculating the density then becomes:

$$\alpha_{ijk} = 1 - exp(-ReLU(\text{raw}_{density,ijk}))$$

The final image pixel value (image $\text{pixel}_{ij}$) is the sum of the product of visibility and density over all M layers:

$$\text{image pixel}_{ij} = \sum_{k=1}^{M} \sigma_{ijk} \alpha_{ijk}$$

Further, incorporating the function T(t) (representing the cumulative transmittance along a ray, i.e., the probability of a ray traveling from camera to the background without encountering any particle). we first calculate the distances ($dists_{ij}$) between the sampled points along the ray. $x_{vals}$ is the distance between the camera to the query points which is on x axis and ray_d as the norm of the ray direction. Then the density ($\alpha_{ijk}$) is calculated as:

$$\alpha_{ijk} = 1 - exp(-ReLU(raw_{density,ijk})(x_{vals,ij(k+1)} - x_{vals,ijk})ray\_d_{ij}$$

Similar to the VR method, the final image pixel value is computed as the sum of the product of visibility and density over all M layers. As depicted in Extended Data Fig. 2, both baselines produced only black images, underscoring that mere volume rendering and the accumulated transmittance function did not yield better results in this problem.

**Model Architecture**
The Free Form Kinematic Self-Model is composite of three integral deep neural networks: the Coordinates Encoder, the Kinematic Encoder, and the Predictive Module. We implement them in Pytorch [36]. We use standard fully connected layers and use ReLU activation [37]. Before we input the coordinates and the joint angles into both encoders, we use Positional Encoding that enables the model to learn higher dimension data. Such an approach has been demonstrated to enhance the performance of Multi-Layer Perceptions [29, 38]. In our work, we use 5 frequencies, so the input dimension of Coordinates Encoder is mapped from 3 to 33 and that of Kinematic Encoder is mapped from 2 to 22. A detailed depiction of this architecture is presented in Extended Data Fig. 3.

**Resolving Initial Training Challenges in Model Training**
During the initial training phase of our model, the model tended to predict an entire image as black (zero), subsequently settling into this as a local optimum. This phenomenon can be attributed to the predominance of black pixels in the image, skewing the model's learning process. In image recognition tasks, such a scenario is often indicative of class imbalance, where one class (in this case, black pixels) dominates the training data, leading the model to a biased learning outcome.

Class imbalance is a well-documented issue in machine learning, particularly in supervised learning tasks. Models tend to favor the majority class, resulting in poor predictive performance for the minority class. This is especially problematic in fields like medical imaging or anomaly detection, where the minority class is often the most critical. Techniques like resampling the dataset, using different initial weights, or applying class-weighted or focal loss functions are commonly employed to address this imbalance [39] [40].

To mitigate this issue in our model, we implemented two strategies:

1. Early abort the training process: If the model only predicts black images, indicating a stagnation in learning, the training process is restarted. This approach helps in escaping from the local optimum that the model initially settles into.
2. Focusing on the central image region: During the first 200 training iterations, we concentrate on the middle section of the image, specifically a 50x50 pixel area. This method reduces the impact of the overwhelming black pixels in the image periphery, allowing the model to better learn features from the more balanced central region.

By adapting these techniques, we aimed to provide the model with a more balanced view of the training data.

**Generation of Spiral Trajectory for Motion Planning**
A spiral trajectory is geometrically complex and requires precise control over the robot's movements. Successfully navigating this trajectory demonstrates the system's ability to handle intricate paths and fine adjustments, showcasing the precision and accuracy of the motion planning algorithms. This section describes the details of generating a spiral trajectory for the motion planning task. The trajectory is composed of three-dimensional coordinates and contains a sequence of 1000 points. The trajectory is formulated by systematically calculating the coordinates of each point along a spiral path. The algorithm operates over a linearly spaced sequence of parameters, ranging from 0 to $4\pi$. For each parameter value in this sequence, the corresponding three-dimensional coordinates of the spiral are computed as follows:

1. The x-coordinate is determined by a linear function of the parameter, signifying a constant rate of increase along the x-axis. t is the parameter varying from 0 to $4\pi$. The x-coordinate for each point is given by the equation:

$$x = 0.01 \times t + 0.02$$

2. The y-coordinate is calculated using a sinusoidal function of the parameter, indicative of a circular motion in the yz-plane. The equation for the y-coordinate is:

$$y = 0.008 \times t \times sin(t)$$

3. Similarly, the z-coordinate is also derived using a cosine function, complementing the y-coordinate to create the spiral effect. The z-coordinate is given by:

$$y = 0.008 \times t \times cos(t) + 0.08$$

These calculations result in a spiral trajectory where each point is defined by its unique set of (x, y, z) coordinates. The spiral progressively extends in the x-direction while exhibiting a circular motion in the yz-plane. The generated trajectory is visualized in Extended Data Fig. 4.

**Utilizing FFKSM for Joint Angle Optimization in Motion Planning Tasks**
In our study, we employ the trained Free Form Kinematic Self-Model (FFKSM) alongside an optimization method to calculate motor commands. This approach enables the robot to accurately reach the target positions along a predefined trajectory, such as a spiral path. Initially, we load this trajectory, which represents a sequence of target points that the robot is tasked to follow. The core objective is to resolve the motor commands using trained FFKSM, facilitating precise navigation to these points.

The parameters for the motion planning algorithm include a loss threshold, set at 10e-5, and a maximum limit of 1000 iterations for the optimization process. The robot's initial joint angles are converted into a tensor format, which facilitates the use of gradient-based optimization techniques. The Adam optimizer is employed for this purpose, known for its effectiveness in handling non-linear optimization problems in robotic control systems.

At the core of the algorithm is an iterative loop, designed to adjust the robot's joint angles and minimize the distance between the predicted position of its end-effector and the target positions along the trajectory. In each iteration, the algorithm resets the optimizer's gradients, and employs FFKSM to predict the end-effector's position. The discrepancy between this predicted position and the target is quantified using a Mean Squared Error (MSE) loss function.

When convergence of the loss below the defined threshold the optimization loop exists. After resolving the motor commands for the entire trajectory. The optimized joint angles corresponding to each target position are saved, allowing for real world robot control to perform manipulation tasks.

During the optimization, if multiple joint configurations can satisfy a particular reference point, the algorithm chooses the configuration that minimizes the overall movement from the previous configuration, promoting smoother transitions and reducing the mechanical wear and energy consumption of the robot.

**Data Availability**

All data for reproduce this work and evaluations are available in the main text or the supplementary materials.

**Code Availability**

The codebase and trained model can be found at https://github.com/H-Y-H-Y-H/fully_body_VSM (DOI: 10.5281/zenodo.11396607) [41]

**Acknowledgements:**

This work was supported in part by the US National Science Foundation (NSF) AI Institute for Dynamical Systems (DynamicsAI.org), grant 2112085

**Author Contributions Statement**

**Yuhang Hu**: Methodology (lead); Writing-original draft (lead); formal analysis (lead); Software (lead); Writing-review & editing (supporting); Conceptualization (equal); **Jiong Lin:** Methodology (supporting); Software (supporting); **Hod Lipson:** Writing-review & editing (lead); Conceptualization (equal); formal analysis (supporting)

**Ethics declarations & Competing Interests Statement**

All the authors conducted this research responsibly and ethically under an inclusive research environment. All the authors declare that they have no competing interests.

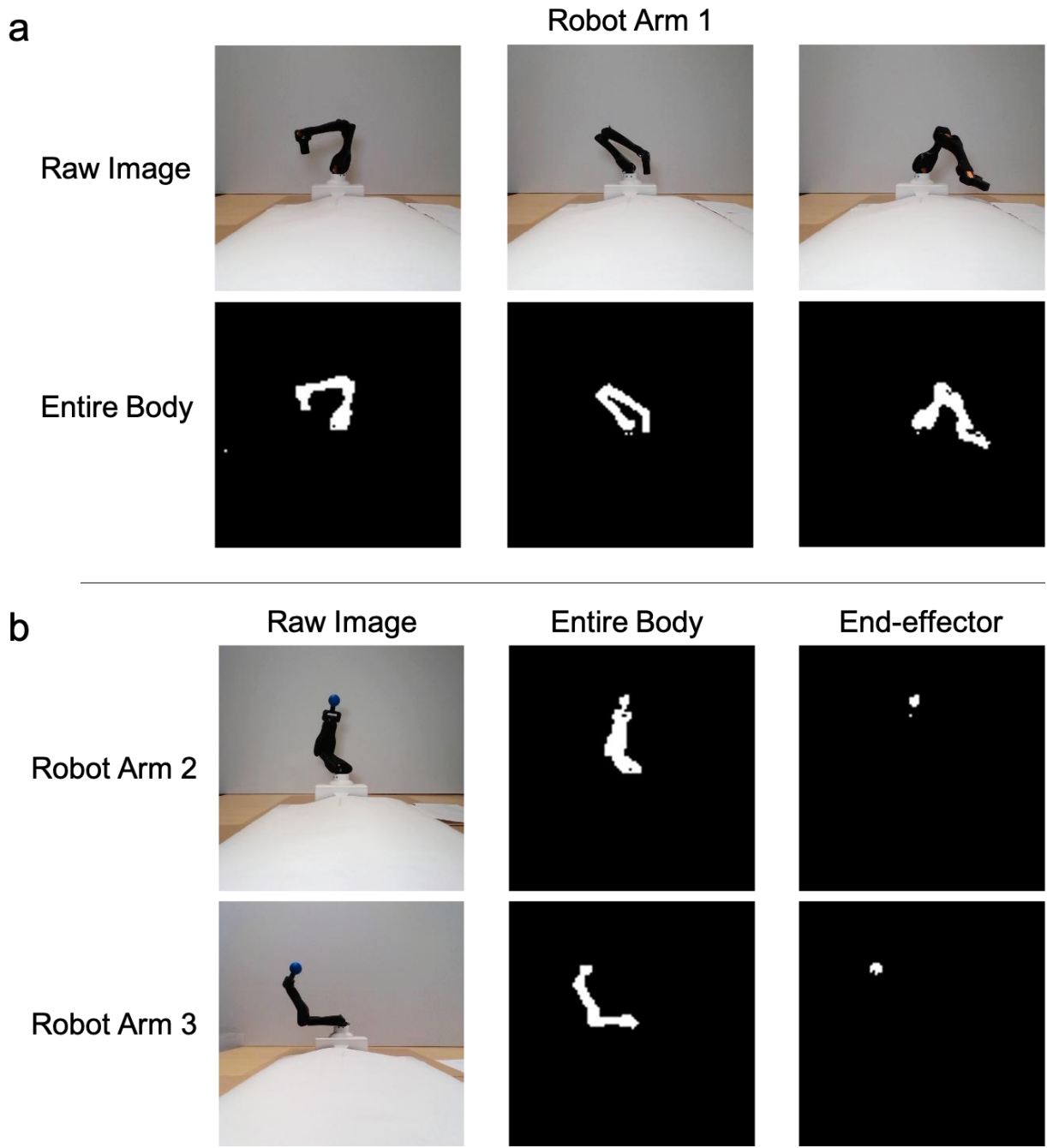

**Extended Data Fig. 1: Data Collection and Segmentation for training FFKSM.** (a) Examples of raw image data alongside their processed binary images. The raw images depict Robot Arm 1 in three distinct positions within the camera's field of view. The corresponding binary images result from color-based segmentation. (b) Example data of Robot Arm 1 and Robot Arm 2. The left side displays raw data images, capturing the robots in different joint positions. The right side shows processed binary images demonstrating the effective isolation of the robot's components (end-effector) through segmentation method we proposed.

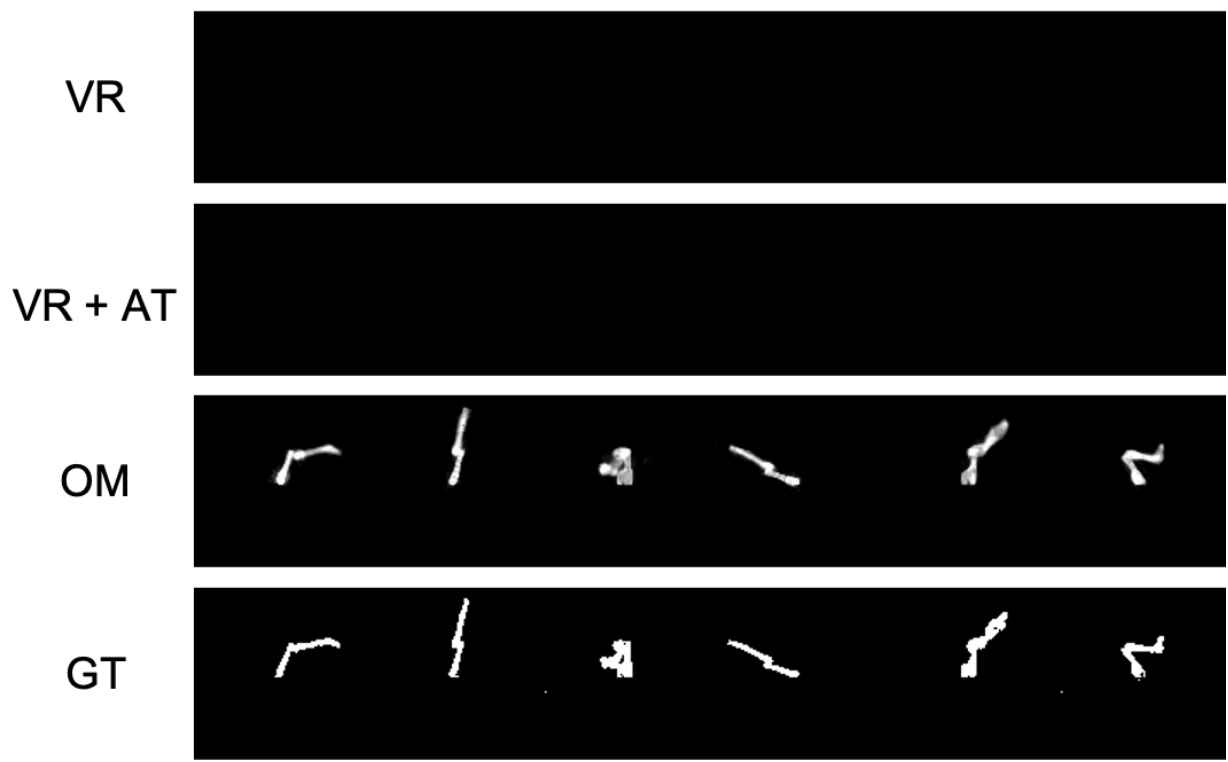

**Extended Data Fig. 2: Comparative Visualization of Different Methods versus Ground Truth**. From top to bottom: (1) Volume Rendering method (VR) - yields a black image indicating ineffective learning, (2) Volume Rendering method with accumulated transmittance function (VR+AT) - similarly results in a black image, (3) Our method (OM) - shows accurate predictions, and (4) Ground truth (GT) for reference.

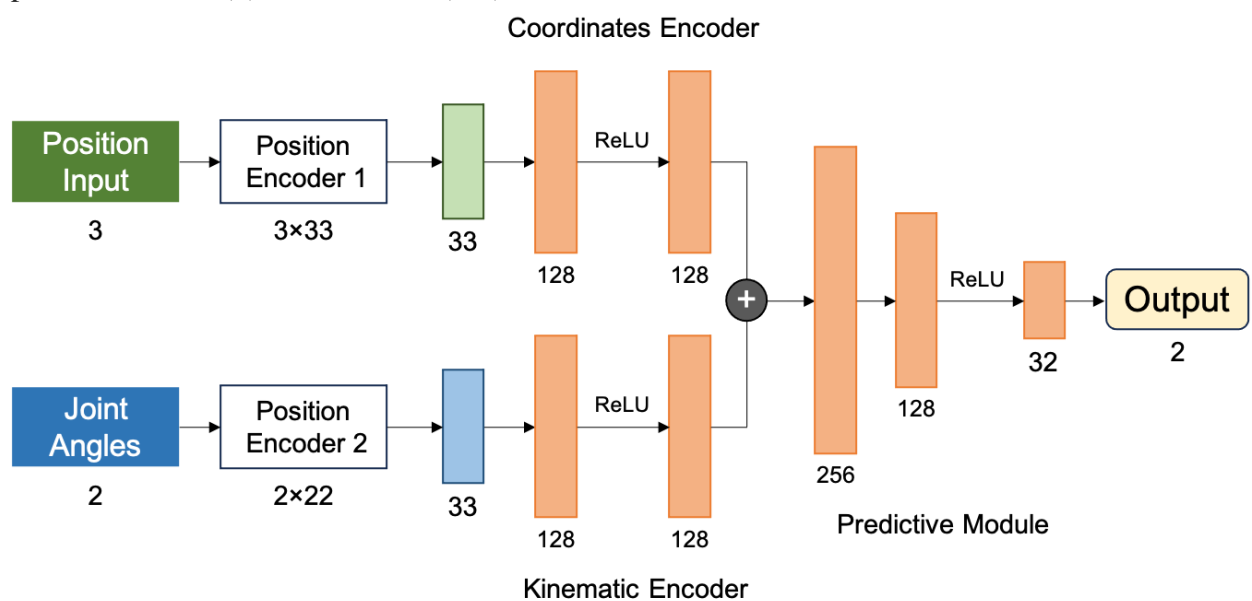

**Extended Data Fig. 3: Architecture of the Free Form Kinematic Self-Model.** This figure illustrates the multi-layer architecture of the Free Form Kinematic Self-Model. Inputs are divided into position and joint angles, each encoded through separate pathways. Positional Encoding enhances the inputs, expanding the 3-dimensional position input to a 33-dimensional vector and the 2-dimensional joint angle input to a 22-dimensional vector. These enhanced inputs are then processed through their respective encoders. The outputs are transformed by ReLU-activated fully connected layers, yielding 128-dimensional feature representations. A subsequent concatenation step combines these features into a 256-dimensional vector, which is further distilled through additional ReLU-activated layers, outputting density and visibility of the input point position.

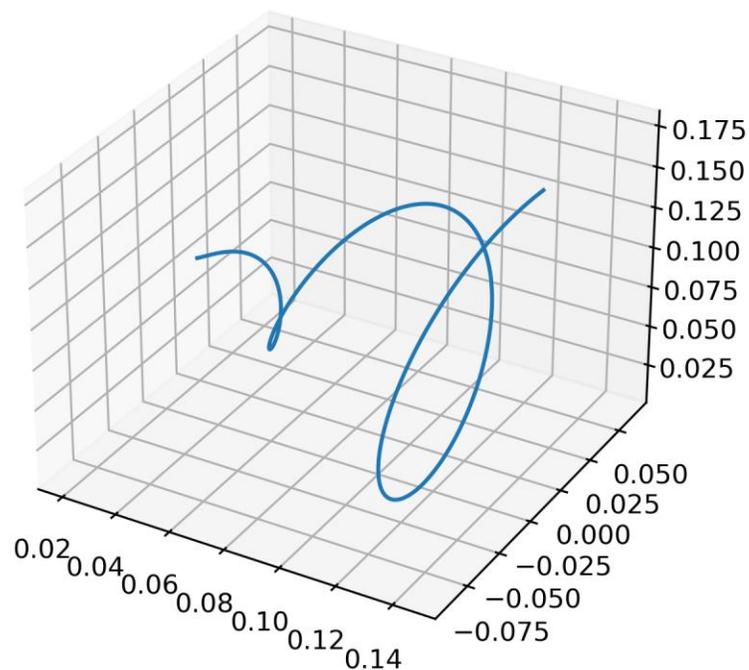

**Extended Data Fig. 4: 3D Visualization of the Spiral Trajectory.** This figure illustrates the spiral trajectory. The spiral starts from the center, gradually extending outward and upward.

**REFERENCES**

[1] A. Afzal, D. S. Katz, C. Le Goues and C. S. Timperley, "Simulation for robotics test automation: Developer perspectives," in *2021 14th IEEE Conference on Software Testing, Verification and Validation (ICST)*, 2021.

[2] H. Choi, C. Crump, C. Duriez, A. Elmquist, G. Hager, D. Han, F. Hearl, J. Hodgins, A. Jain, F. Leve and others, "On the use of simulation in robotics: Opportunities, challenges, and suggestions for moving forward," *Proceedings of the National Academy of Sciences,* vol. 118, p. e1907856118, 2021.

[3] C. K. Liu and D. Negrut, "The role of physics-based simulators in robotics," *Annual Review of Control, Robotics, and Autonomous Systems,* vol. 4, p. 35–58, 2021.

[4] L. Žlajpah, "Simulation in robotics," *Mathematics and Computers in Simulation,* vol. 79, p. 879–897, 2008.

[5] D. Howard, A. E. Eiben, D. F. Kennedy, J.-B. Mouret, P. Valencia and D. Winkler, "Evolving embodied intelligence from materials to machines," *Nature Machine Intelligence,* vol. 1, p. 12–19, 2019.

[6] G. G. Gallup, "Self recognition in primates: A comparative approach to the bidirectional properties of consciousness.," *American psychologist,* vol. 32, p. 329, 1977.

[7] G. G. Gallup Jr, J. R. Anderson and D. J. Shillito, "The mirror test," *The cognitive animal: Empirical and theoretical perspectives on animal cognition,* p. 325–333, 2002.

[8] G. G. Gallup Jr, "Self-awareness and the emergence of mind in primates," *American Journal of Primatology,* vol. 2, p. 237–248, 1982.

[9] T. F. Cash, Body image., Oxford University Press, 2000.

[10] H. J. Chiel and R. D. Beer, "The brain has a body: adaptive behavior emerges from interactions of nervous system, body and environment," *Trends in neurosciences,* vol. 20, p. 553–557, 1997.

[11] W. Agnew, C. Xie, A. Walsman, O. Murad, Y. Wang, P. Domingos and S. Srinivasa, "Amodal 3d reconstruction for robotic manipulation via stability and connectivity," in *Conference on Robot Learning*, 2021.

[12] W. Huang, C. Wang, R. Zhang, Y. Li, J. Wu and L. Fei-Fei, "Voxposer: Composable 3d value maps for robotic manipulation with language models," *arXiv preprint arXiv:2307.05973,* 2023.

[13] C. Papachristos, S. Khattak, F. Mascarich, T. Dang and K. Alexis, "Autonomous aerial robotic exploration of subterranean environments relying on morphology–aware path planning," in *2019 International Conference on Unmanned Aircraft Systems (ICUAS)*, 2019.

[14] Z. Xu, Z. He, J. Wu and S. Song, "Learning 3d dynamic scene representations for robot manipulation," *arXiv preprint arXiv:2011.01968,* 2020.

[15] L. Steels and M. Spranger, "The robot in the mirror," *Connection Science,* vol. 20, p. 337–358, 2008.

[16] R. Kwiatkowski and H. Lipson, "Task-agnostic self-modeling machines," *Science Robotics,* vol. 4, p. eaau9354, 2019.

[17] R. Kwiatkowski, Y. Hu, B. Chen and H. Lipson, "On the Origins of Self-Modeling," *arXiv preprint arXiv:2209.02010,* 2022.

[18] R. Vaughan and M. Zuluaga, "Use your illusion: Sensorimotor self-simulation allows complex agents to plan with incomplete self-knowledge," in *International Conference on Simulation of Adaptive Behavior*, 2006.

[19] S. Wittmeier, C. Alessandro, N. Bascarevic, K. Dalamagkidis, D. Devereux, A. Diamond, M. Jäntsch, K. Jovanovic, R. Knight, H. G. Marques and others, "Toward anthropomimetic robotics: development, simulation, and control of a musculoskeletal torso," *Artificial life,* vol. 19, p. 171–193, 2013.

[20] C. Blum, A. F. T. Winfield and V. V. Hafner, "Simulation-based internal models for safer robots," *Frontiers in Robotics and AI,* vol. 4, p. 74, 2018.

[21] B. Chen, R. Kwiatkowski, C. Vondrick and H. Lipson, "Fully body visual self-modeling of robot morphologies," *Science Robotics,* vol. 7, p. eabn1944, 2022.

[22] J. T. Barron, B. Mildenhall, M. Tancik, P. Hedman, R. Martin-Brualla and P. P. Srinivasan, "Mip-nerf: A multiscale representation for anti-aliasing neural radiance fields," in *Proceedings of the IEEE/CVF International Conference on Computer Vision*, 2021.

[23] A. Pumarola, E. Corona, G. Pons-Moll and F. Moreno-Noguer, "D-nerf: Neural radiance fields for dynamic scenes," in *Proceedings of the IEEE/CVF Conference on Computer Vision and Pattern Recognition*, 2021.

[24] C. Reiser, S. Peng, Y. Liao and A. Geiger, "Kilonerf: Speeding up neural radiance fields with thousands of tiny mlps," in *Proceedings of the IEEE/CVF International Conference on Computer Vision*, 2021.

[25] Mildenhall, B. et al. NeRF: representing scenes as neural radiance fields for view synthesis. *Commun. ACM* **65**, 99–106 (2021).

[26] B. Hu, J. Huang, Y. Liu, Y.-W. Tai and C.-K. Tang, "NeRF-RPN: A general framework for object detection in NeRFs," in *Proceedings of the IEEE/CVF Conference on Computer Vision and Pattern Recognition*, 2023.

[27] V. Lazova, V. Guzov, K. Olszewski, S. Tulyakov and G. Pons-Moll, "Control-nerf: Editable feature volumes for scene rendering and manipulation," in *Proceedings of the IEEE/CVF Winter Conference on Applications of Computer Vision*, 2023.

[28] C. Xu, B. Wu, J. Hou, S. Tsai, R. Li, J. Wang, W. Zhan, Z. He, P. Vajda, K. Keutzer and others, "Nerf-det: Learning geometry-aware volumetric representation for multi-view 3d object detection," in *Proceedings of the IEEE/CVF International Conference on Computer Vision*, 2023.

[29] E. D. Zhong, T. Bepler, J. H. Davis and B. Berger, "Reconstructing continuous distributions of 3D protein structure from cryo-EM images," *arXiv preprint arXiv:1909.05215,* 2019.

[30] J. Xu, L. Peng, H. Cheng, H. Li, W. Qian, K. Li, W. Wang and D. Cai, "Mononerd: Nerf-like representations for monocular 3d object detection," in *Proceedings of the IEEE/CVF International Conference on Computer Vision*, 2023.

[31] D. P. Kingma and J. Ba, "Adam: A method for stochastic optimization," *arXiv preprint arXiv:1412.6980,* 2014.

[32] S. LaValle, "Rapidly-exploring random trees: A new tool for path planning," *Research Report 9811,* 1998.

[33] J. Bongard, V. Zykov and H. Lipson, "Resilient machines through continuous self-modeling," *Science,* vol. 314, p. 1118–1121, 2006.

[34] S. Kucuk and Z. Bingul, Robot kinematics: Forward and inverse kinematics, INTECH Open Access Publisher London, UK, 2006.

[35] E. Coumans and Y. Bai, "Pybullet, a python module for physics simulation for games, robotics and machine learning," 2016.

[36] A. Paszke, S. Gross, F. Massa, A. Lerer, J. Bradbury, G. Chanan, T. Killeen, Z. Lin, N. Gimelshein, L. Antiga and others, "Pytorch: An imperative style, high-performance deep learning library," *Advances in neural information processing systems,* vol. 32, 2019.

[37] A. F. Agarap, "Deep learning using rectified linear units (relu)," *arXiv preprint arXiv:1803.08375,* 2018.

[38] M. Tancik, P. Srinivasan, B. Mildenhall, S. Fridovich-Keil, N. Raghavan, U. Singhal, R. Ramamoorthi, J. Barron and R. Ng, "Fourier features let networks learn high frequency functions in low dimensional domains," *Advances in Neural Information Processing Systems,* vol. 33, p. 7537–7547, 2020.

[39] M. Buda, A. Maki and M. A. Mazurowski, "A systematic study of the class imbalance problem in convolutional neural networks," *Neural networks,* vol. 106, p. 249–259, 2018.

[40] T.-Y. Lin, P. Goyal, R. Girshick, K. He and P. Dollár, "Focal loss for dense object detection," in *Proceedings of the IEEE international conference on computer vision*, 2017.

[41] H-Y-H-Y-H, jiong lin, & ArthasL1. (2024). H-Y-H-Y-H/fully_body_VSM: publication (publication). Zenodo. https://doi.org/10.5281/zenodo.11396607